\documentclass[letterpaper]{article}
\pdfoutput=1
\usepackage{aaai25}  
\usepackage{times}  
\usepackage{helvet}  
\usepackage{courier}  
\usepackage[hyphens]{url}  
\usepackage{graphicx} 
\usepackage{natbib}  
\usepackage{caption} 
\usepackage{algorithm}
\usepackage{algorithmicx}
\usepackage{algpseudocode}
\usepackage{amsmath}
\usepackage{xcolor}
\usepackage{amsfonts} 

\usepackage{newfloat}
\usepackage{listings}
\DeclareCaptionStyle{ruled}{labelfont=normalfont,labelsep=colon,strut=off} 
\floatstyle{ruled}
\newfloat{listing}{tb}{lst}{}
\floatname{listing}{Listing}

\title{Perceptual Equilibrium: A Game Theoretical-based}
\title{SPRIG: Stackelberg Perception-Reinforcement Learning with Internal Game Dynamics}
\author {
    Fernando Martinez-Lopez\textsuperscript{\rm 1},
    Juntao Chen\textsuperscript{\rm 1},
    Yingdong Lu\textsuperscript{\rm 2}
}
\affiliations {
    \textsuperscript{\rm 1}Fordham University\\
    \textsuperscript{\rm 2}IBM Research\\
    \{fmartinezlopez, jchen504\}@fordham.edu, yingdong@us.ibm.com
}

\usepackage{bibentry}

\begin{document}

\maketitle

\begin{abstract}

Deep reinforcement learning agents often face challenges to effectively coordinate perception and decision-making components, particularly in environments with high-dimensional sensory inputs where feature relevance varies. This work introduces SPRIG (Stackelberg Perception-Reinforcement learning with Internal Game dynamics), a framework that models the internal perception-policy interaction within a single agent as a cooperative Stackelberg game. In SPRIG, the perception module acts as a leader, strategically processing raw sensory states, while the policy module follows, making decisions based on extracted features. SPRIG provides theoretical guarantees through a modified Bellman operator while preserving the benefits of modern policy optimization. Experimental results on the Atari BeamRider environment demonstrate SPRIG's effectiveness, achieving around 30\% higher returns than standard PPO through its game-theoretical balance of feature extraction and decision-making.
\end{abstract}

\section{Introduction}
Deep Reinforcement Learning (RL) has successfully solved complex tasks across various domains, from game playing to robotic control \cite{mnih2015human, berner2019dota, Ibarz2021HowTT}. However, a fundamental challenge persists: the effective coordination between perception and decision-making components, particularly in environments with high-dimensional sensory inputs where the relevance of features varies across tasks or time \cite{mao2024pdit}. 

While traditional approaches treat perception and decision-making as a unified process, this integration overlooks fundamental insights from cognitive science, particularly the two-stream hypothesis of visual processing \cite{goodale1992separate}. In biological systems, visual information flows through distinct pathways: a ``what" stream for object recognition and a ``how" stream for action guidance. This natural division suggests an inherent hierarchy where perceptual processing precedes and informs action selection as separate subsystems. Despite this biological inspiration, current RL approaches lack the fundamentals of this natural cooperative design.

\begin{figure}[t]
\centering
\includegraphics[trim={7cm 3cm 7cm 4cm}, clip, width=1\columnwidth]{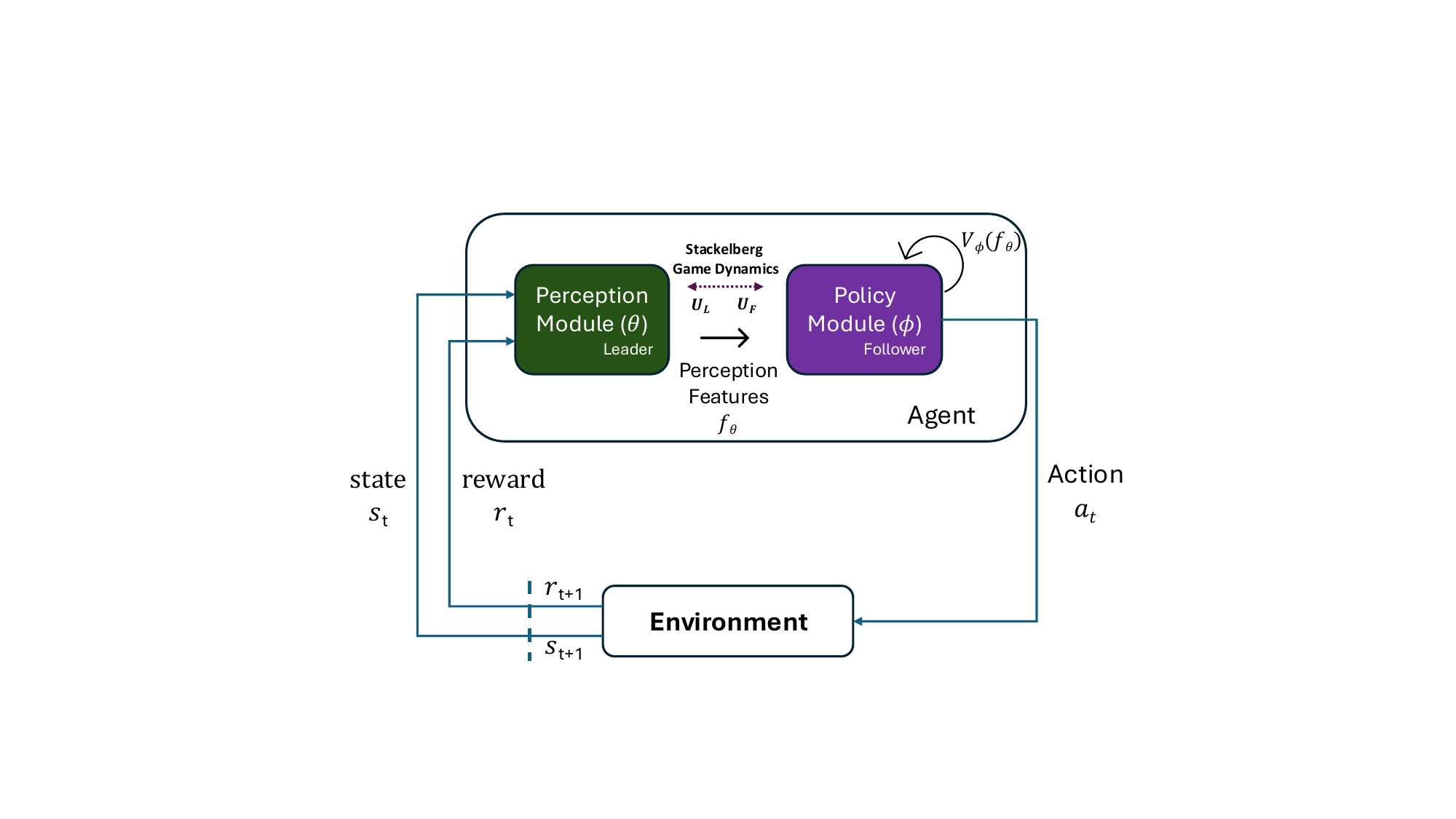}
\caption{SPRIG architecture overview}
\label{fig:sprig_overview}
\end{figure}

We address these challenges by introducing SPRIG (Stackelberg Perception-Reinforcement learning with Internal Game dynamics), a framework that models the perception-policy interaction as a cooperative Stackelberg game. Our \emph{perception module}, implemented as a hierarchical spatio-temporal attention mechanism, acts as a leader that strategically processes raw sensory inputs, while the \emph{policy module} follows by making decisions based on the extracted features. Our game-theoretical formulation provides: (1) a principled mathematical framework for perception-policy interaction through a modified Bellman operator, (2) thorough theoretical analysis with provable convergence properties while maintaining the advantages of modern policy optimization, and (3) the creation of a natural balance between feature extraction and policy optimization through our cooperative formulation. Our approach extends the Proximal Policy Optimization algorithm (PPO; \cite{schulman2017proximalpolicyoptimizationalgorithms}) to incorporate this game-theoretical dynamic, introducing a two-stage optimization process with advantage normalization. Through our formulation of perception cost and utility functions, we ensure that our method converges to a unique fixed point, providing both theoretical soundness and practical applicability. 

We demonstrate the effectiveness of SPRIG on the Atari BeamRider environment, where the perception module must identify and track relevant visual features for successful policy learning. Our empirical results show that SPRIG achieves higher than standard PPO, with returns reaching approximately 850 versus 650 for the baseline. 

\section{Related Work}
Integrating perception and decision-making in RL has become an interesting subdomain, especially in environments with high-dimensional sensory inputs. The Perception and Decision-making Interleaving Transformer (PDiT; \cite{mao2024pdit}) uses separate transformers for perception and decision-making, leading to enhanced performance in complex tasks. Incorporating game-theoretic principles, \cite{zheng2022stackelberg} proposed the Stackelberg Actor-Critic framework, modeling the actor-critic interaction as a Stackelberg game to improve learning stability. Extending this approach, \cite{huangrobust2022} addressed robustness in uncertain environments by formulating robust RL as a Stackelberg game, demonstrating the adaptability of leader-follower structures in RL. Attention mechanisms have also been explored for adaptive feature extraction in RL. For instance, \cite{manchin2019reinforcement} introduced a self-supervised attention model that significantly improved performance in the Arcade Learning Environment, highlighting the potential of attention mechanisms in RL.

Our work advances these approaches by introducing a framework with theoretical guarantees through a modified Bellman operator that explicitly accounts for perception-policy interaction, while maintaining the advantages of modern policy optimization. Our cooperative game formulation creates a natural balance between feature extraction and decision-making, complementing previous approaches by adding provable convergence properties for the entire system and demonstrating empirical improvements.

\section{Background and Preliminaries}

\subsection{Markov Decision Processes and Reinforcement Learning}
A Markov Decision Process (MDP) provides the fundamental model for sequential decision-making under uncertainty \cite{sutton2018reinforcement}. Formally, an MDP is defined as a tuple \(\mathcal{M} = (S, A, P, R, \gamma)\), where \(S\) represents the state space, \(A\) the action space, \(P: S \times A \times S \rightarrow [0,1]\) the transition probability function, \(R: S \times A \rightarrow \mathbb{R}\) the reward function, and \(\gamma \in [0,1)\) the discount factor.

Here an agent interacts with the environment by selecting actions according to a policy \(\pi: S \rightarrow \Delta(A)\), where \(\Delta(A)\) denotes the probability simplex over actions. The objective is to find an optimal policy \(\pi^*\) that maximizes the expected discounted return:
\begin{equation}
    V^\pi(s) = \mathbb{E}_{\pi}\left[\sum_{t=0}^{\infty} \gamma^t R(s_t, a_t) \mid s_0 = s\right].
\end{equation}


The optimal policy $\pi^*$ satisfies the Bellman optimality equation:
\begin{equation}
V^*(s) = \max_{a \in A} \left[R(s,a) + \gamma \mathbb{E}_{s' \sim P(\cdot|s,a)}[V^*(s')]\right].    
\end{equation}

\subsection{Stackelberg and Cooperative Games}
Stackelberg games model sequential decision-making scenarios through a hierarchical structure. Let \(\mathcal{G} = (N, \Theta, \Phi, u_L, u_F)\) be a two-player game where \(N = \{L,F\}\) denotes the leader and follower, with strategy spaces \(\Theta\) and \(\Phi\) respectively. The utility functions \(u_L: \Theta \times \Phi \rightarrow \mathbb{R}\) and \(u_F: \Theta \times \Phi \rightarrow \mathbb{R}\) define the payoffs for each player, though their usage differs due to the sequential nature of the game.

In this interplay, the leader commits to a strategy \(\theta \in \Theta\), after which the follower observes this commitment and responds with \(\phi \in \Phi\). This creates a subgame perfect equilibrium where the follower's best response function is:
\begin{equation}
 BR_F(\theta) = \{\phi \in \Phi : u_F(\theta, \phi) \geq u_F(\theta, \phi') \text{ for all } \phi' \in \Phi\}.   
\end{equation}
The leader, anticipating this response, solves:
\begin{equation}
\theta^* = \arg\max_{\theta \in \Theta} u_L(\theta, BR_F(\theta)).   
\end{equation}

While Stackelberg games capture hierarchical interaction, cooperative game theory provides tools for analyzing scenarios where players coordinate for mutual benefit. A cooperative game is defined by \((N,v)\), where \(N\) is the player set and \(v: 2^N \rightarrow \mathbb{R}\) is the characteristic function assigning values to coalitions.

In our two-player setting, the cooperative value emerges through a weighted combination of individual utilities:
\begin{equation}
    v(\{L,F\}) = \alpha u_L(\theta, \phi) + (1-\alpha) u_F(\theta, \phi),
\end{equation}
where \(\alpha \in [0,1]\) represents the cooperation weight. The solution concept focuses on finding allocations that maximize this joint value while ensuring individual rationality: \(v(\{i\}) \leq u_i\) for \(i \in \{L,F\}\).

\subsection{Perception-Policy Learning: Motivation and Need}
Current approaches to perception-policy learning typically fall into two categories. The first approach treats perception and policy as a single end-to-end system, while the second attempts to optimize these components independently. Nevertheless, recent work has shown that perception and decision models separately can lead to reduce robustness since mismatched state extraction and control decision-making become asynchronous \cite{zhuAutonomous}. Standard RL models often treat these processes as a unified pipeline, optimizing perception and policy jointly in an end-to-end fashion. While this approach simplifies implementation, it struggles to generalize in high-dimensional environments where irrelevant features dominate or feature relevance varies over time, as evidenced in complex visual navigation tasks \cite{zhu2017target}. Such limitations derives from the inability to effectively balance the demands of feature extraction with those of action selection.

Inspired by the two-stream hypothesis of visual processing \cite{goodale1992separate}, we argue for a principled separation of perception and policy into distinct, hierarchically organized modules, aligning with approaches in hierarchical RL that decompose complex tasks \cite{diuk2013divide}. This biological insight suggests that perception should focus on extracting meaningful, task-relevant features while policy concentrates on optimal action selection based on these features. This separation enables better modularity and adaptability in complex environments, akin to how biological systems achieve robust and efficient decision-making.

However, decoupling perception from policy introduces coordination challenges. Misalignment between the extracted features and the policy's decision-making objectives can degrade performance, necessitating a structured design to guide this interaction. Game theory, particularly Stackelberg games, provides a natural solution. By modeling the perception module as a leader and the policy module as a follower, we establish a hierarchical interaction where the perception module optimizes its feature extraction strategy while anticipating the policy module's response.

Through this hierarchical formulation, we address the shortcomings of traditional RL methods, introducing a modular, biologically inspired framework capable of robust generalization in complex tasks. This structured interaction also facilitates theoretical analysis and practical improvements, setting the foundation for the novel game-theoretical approach introduced in this paper.

\section{Our Approach}
We propose SPRIG (Stackelberg Perception-Reinforcement Learning with Internal Game Dynamics), a cooperative Stackelberg game framework for perception-policy learning in RL. Our approach builds upon PPO, extending it to incorporate game-theoretical dynamics between modules. As shown in Figure.~\ref{fig:sprig_overview}, the SPRIG architecture comprises two key components: the \emph{perception module}, which acts as the leader, and the \emph{policy module}, which serves as the follower.

\subsection{Perception-Policy Game Formulation}
In our SPRIG agent, the perception module \(\theta\) implements a hierarchical spatio-temporal attention mechanism consisting of three convolutional layers combined with self-attention, mapping raw states \(S\) to features \(\mathcal{F}\). This way, the agent can process raw visual inputs by progressively refining spatial relationships while maintaining temporal consistency across frames. On the other hand, the policy module \(\phi\) consists of a Multi-Layer Perception that takes the feature representation coming from the perception module and outputs action probabilities. The policy module is optimized iteratively using PPO, alternating between policy updates and value function updates.

The interaction between these modules is formulated as a cooperative Stackelberg game where:
\begin{align}
    \theta^* = \arg\max_{\theta \in \Theta} u_L(\theta, \phi^*(\theta)),\\
    \phi^*(\theta) = \arg\max_{\phi \in \Phi} \mathbb{E}_{\pi_\phi}[R(s,a)].
\end{align}
The leader's utility function \(u_L\) balances both the policy's performance and the perception computational efficiency:
\begin{equation}
\label{eqn:perception_optimization}
u_L(\theta, \phi) = \alpha_{coop} \mathbb{E}_{\pi_\phi}[R(s,a)] - (1-\alpha_{coop})C_\theta(s). 
\end{equation}
where \(\alpha_{coop}\) is the cooperation weight and \(C_\theta(s)\) represents the perception cost. The cost function penalizes excessive attention across all layers:
\begin{equation}
\label{eqn:perception_cost}
C_\theta(s) = \lambda_c \sum_{k=1}^K \mathbb{E}_{s \sim \mathcal{D}}[\|A_k(s)\|_1], 
\end{equation}
where \(A_k(s)\) represents the attention weights at layer \(k\), \(\lambda\) is the cost weight, \(\mathcal{D}\) is the distribution of states encountered during training, and \(K\) is the total number of layers.

\subsection{Stackelberg Equilibrium Computation}
The Stackelberg equilibrium is computed through a two-stage optimization process. In the first stage, the perception module optimizes its utility while anticipating the policy module's response (Eq. (\ref{eqn:perception_optimization})):
\begin{equation}
    \mathcal{L}_\theta = -u_L(\theta, \phi). 
\end{equation}
This optimization uses Generalized Advantage Estimation (GAE; \cite{Schulman2015HighDimensionalCC}) to compute advantages, which are normalized for training stability. The perception cost directly influences this stage by penalizing excessive attention allocation.

In the second stage, the policy module optimizes its objective given the features provided by the perception module:
\begin{equation}
    \mathcal{L}_\phi = -\mathbb{E}_{\pi_\phi}[R(s,a)] + \beta H(\pi_\phi),
\end{equation}
where \(H(\pi_\phi)\) is the policy entropy and \(\beta\) is the entropy coefficient. The policy optimization includes both value function and policy updates, with the perception cost indirectly affecting this stage through the quality of extracted features, as outlined in Algorithm~\ref{alg:stackelberg_ppo}.

The perception cost influences both optimization stages: directly in the leader's utility computation and indirectly in the follower's optimization through feature quality. This dual influence creates a balanced cooperation between modules, where the perception module must provide useful features while maintaining computational efficiency, and the policy module must effectively utilize these features for decision-making.

\begin{algorithm}[!t]
\caption{SPRIG Agent: Cooperative Stackelberg Game Training for Perception-Policy Learning}
\label{alg:stackelberg_ppo}
\begin{algorithmic}[1]
\Require Initial parameters $\theta$ for perception, $\phi$ for policy module.
\Require Cooperation weight $\alpha$, discount factor $\gamma$, GAE parameter $\lambda$
\For{each iteration}
    \State Collect trajectories $\mathcal{D}$ using current policy
    \State Compute returns and normalize GAE values  $\hat{A}_t$
    \For{each PPO epoch}
        \For{each mini-batch $\mathcal{B}$}
            \State // Leader (Perception) Stage
            \State $f_\theta \gets$ perception features for states in $\mathcal{B}$
            \State $C_\theta$ \Comment{Attention cost, Equation \eqref{eqn:perception_cost}}
            \State $\pi_\phi \gets$ policy distribution from $f_\theta$
            \State $u_{\text{policy}} \gets (\log \pi_\phi(a) \cdot \hat{A}_t)_{\text{mean}}$
            \State $u_L \gets \alpha_{coop}(-C_\theta) + (1-\alpha_{coop})u_{\text{policy}}$
            \State Update $\theta$ by maximizing $u_L$ with gradient clipping
            
            \State // Follower (Policy) Stage
            \State Compute PPO ratio $r_t(\phi)$
            \State $\mathcal{L}_{\text{CLIP}} \gets \min(r_t(\phi)\hat{A}_t, \text{clip}(r_t(\phi), 1\pm\epsilon)\hat{A}_t)$
            \State $\mathcal{L}_V \gets (V_\phi(s_t) - R_t)^2$
            \State $\mathcal{L}_\phi \gets -\mathcal{L}_{\text{CLIP}} + 0.5\mathcal{L}_V - 0.01H(\pi_\phi) + C_\theta$
            \State Update $\phi$ by minimizing $\mathcal{L}_\phi$ with gradient clipping
        \EndFor
    \EndFor
\EndFor
\end{algorithmic}
\end{algorithm}

\subsection{Theoretical Formulation and Convergence Properties}

\subsubsection{Stackelberg-MDP Formulation}
We extend the traditional MDP framework to incorporate the perception-policy interaction through a cooperative Stackelberg game. Our augmented MDP is defined as \(\mathcal{M}_S = (S, A, P, R, \gamma, \Theta, \Phi, C)\), where \(\Theta\) is the perception parameter space, \(\Phi\) is the policy parameter space, and \(C: S \times \Theta \rightarrow [0,1]\) is the perception cost function implemented through attention mechanisms.

\subsubsection{Bellman Operator and Properties}
For our Stackelberg-MDP, we first define the standard Bellman operator \(\mathcal{T}\) for MDPs:
\begin{equation}
   (\mathcal{T}f)(s,a) = R(s,a) + \gamma \mathbb{E}_{s' \sim P(\cdot|s,a)}[\max_{a'} f(s',a')].
\end{equation}

Building upon this, we define our Stackelberg-Bellman operator \(\mathcal{T}_S\) that incorporates the perception-policy interaction:
\begin{equation}
\begin{split}
(\mathcal{T}_S f)(s,a) = &\max_{\theta \in \Theta} \min_{\phi \in \Phi} 
\Big[ R(s,a) - \lambda C_\theta(s) \\
&+ \gamma \mathbb{E}_{s' \sim P(\cdot|s,a)} 
\big[f(s',a'; \phi\big] \Big],
\end{split}
\end{equation}
where \(C_\theta(s) = \sum_{k=1}^K \|A_k(s)\|_1\) represents our implemented attention-based perception cost (Equation \eqref{eqn:perception_cost}).

\subsubsection{Contraction Properties}
The Stackelberg-Bellman operator \(\mathcal{T}_S\) maintains the contraction property under the following conditions: 1) \textit{bounded rewards}: \(|R(s,a)| \leq R_{max}\); 2) \textit{bounded perception cost}: \(0 \leq C_\theta(s) \leq 1\) (guaranteed by our $L_1$-norm attention cost); and 3) \textit{discount factor}: \(\gamma \in [0,1)\). For any two value functions \(f_1\) and \(f_2\):
\begin{equation}
\|\mathcal{T}_S f_1 - \mathcal{T}_S f_2\|_\infty \leq \gamma \|f_1 - f_2\|_\infty.    
\end{equation}
The contraction property of \(\mathcal{T}_S\) ensures the existence of a unique fixed point \(f^*\) satisfying \(f^* = \mathcal{T}_S f^*\), guarantees convergence of value iteration: \(\|\mathcal{T}_S^n f - f^*\|_\infty \leq \gamma^n \|f - f^*\|_\infty\), and that the optimal policy derived from \(f^*\) represents the Stackelberg equilibrium between perception and policy modules.

\section{Numerical Experiments}
\begin{figure}[t]
\centering
\includegraphics[width=1.0\columnwidth]{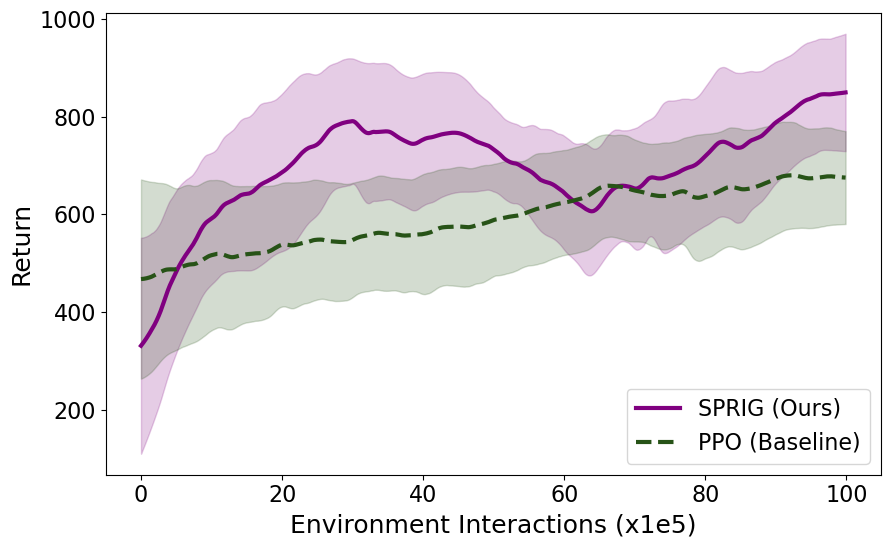} 
\caption{Return curves for SPRIG and baseline PPO on BeamRider. Results averaged across 5 seeds with shaded regions showing standard deviation.}
\label{fig:results}
\end{figure}

We evaluate our SPRIG agent on the BeamRider Atari environment, conducting experiments across five different random seeds over 10 million environment interactions. We considered BeamRider as an interesting challenge for our agent since temporal element and visual focus are important in this game. We utilize identical hyperparameters with both the baseline (PPO) and SPRIG except for the perception module configuration. The detailed perception module architecture specifications are provided in Appendix, Fig.~\ref{fig:perception_module}. SPRIG achieves superior performance compared to the baseline PPO implementation as presented in Fig.~\ref{fig:results}, reaching approximately 850 points compared to PPO's 650, showing a clear advantage in the final performance. The learning process exhibits interesting dynamics: SPRIG demonstrates faster initial learning in the first 2 million steps, followed by a period of exploration and adjustment between 4-6 million steps, before stabilizing at a higher performance level. While the learning trajectory shows higher variance during the middle phase (as indicated by the purple-shaded region), this exploration appears beneficial for discovering better policies, ultimately leading to more robust performance. The baseline PPO, in contrast, shows more stable but conservative learning, with a steady but slower improvement curve and lower final performance. These results suggest that our game-theoretical framework effectively balances the exploration-exploitation trade-off while maintaining learning stability.

\section{Conclusions}
In this paper, we presented SPRIG, a novel framework that formalizes perception-policy interaction in reinforcement learning through cooperative Stackelberg games. Our approach provides theoretical guarantees through a modified Bellman operator while demonstrating practical improvements in learning efficiency and stability. The preliminary results suggest that explicitly modeling module interaction through game theory could be a promising direction for improving single-agent reinforcement learning systems.

\bibliography{aaai25}

\section{Appendix}

\begin{table}[h]
\centering
\begin{tabular}{l|r}
\hline
Rollout Length & 2048 \\
Batch Size & 64 \\
Discount Factor ($\gamma$) & 0.99 \\
GAE Parameter ($\lambda$) & 0.95 \\
Learning Rate & 1e-4 \\
PPO Epochs & 4 \\
PPO Clip Range ($\epsilon$) & 0.2 \\
Value Coefficient & 0.5 \\
Entropy Coefficient & 0.01 \\
Max Grad Norm (gradient clipping) & 0.5 \\
Perception Cost Weight ($\lambda_c$) & 1e-4 \\
Cooperation Weight ($\alpha_{coop}$) & 0.7 \\
Total Timesteps & 1e7 \\
Max Episode Length & 10000 \\
\hline
\end{tabular}
\caption{Hyperparameters for SPRIG \& PPO}
\label{appendix_hyperparameters}
\end{table}

\begin{figure}[t]
\centering
\includegraphics[trim={7cm 3cm 2cm 0cm}, clip, width=1.2\columnwidth]{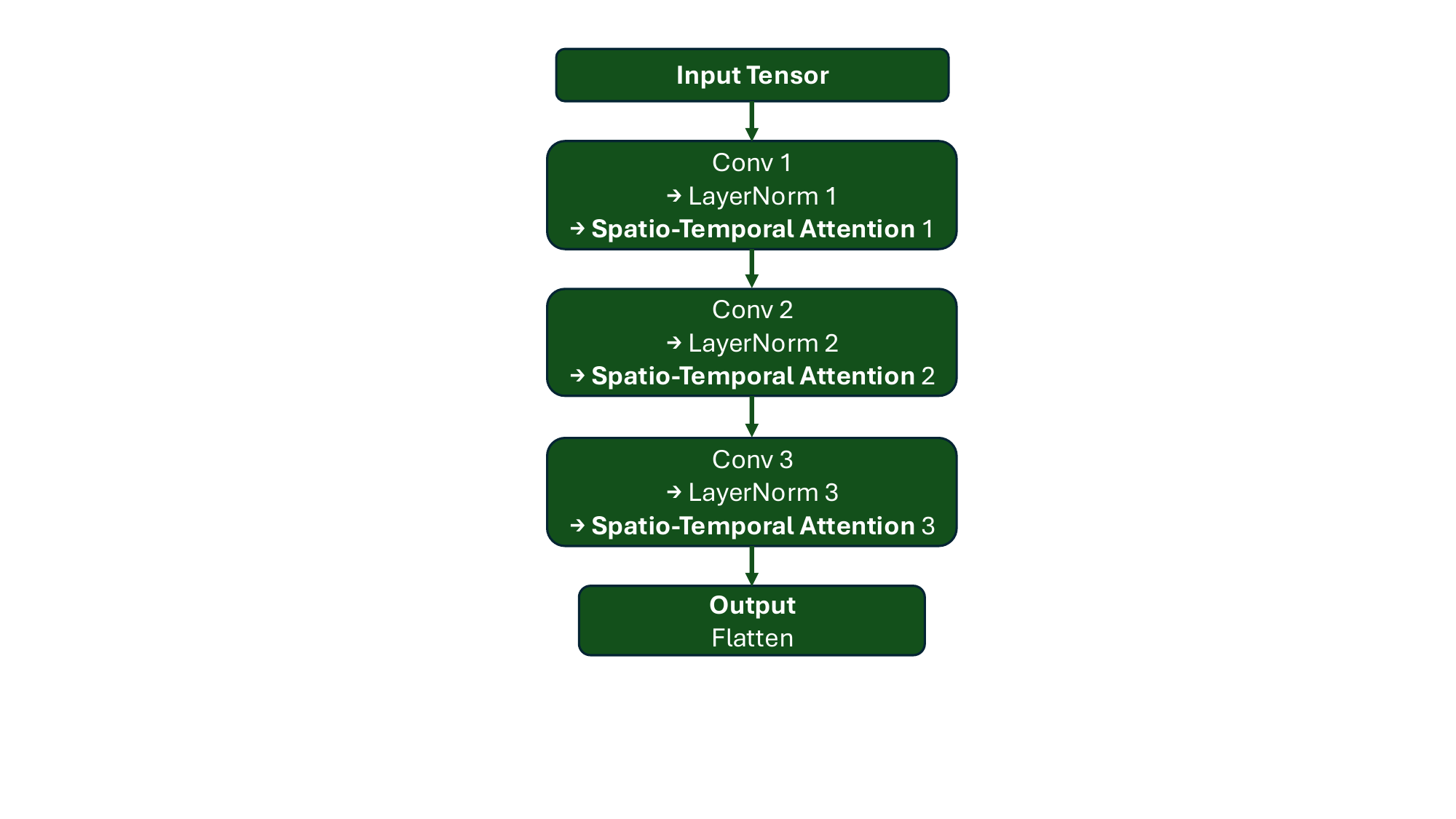}
\caption{Perception Module ($\theta$)}
\label{fig:perception_module}
\end{figure}

\begin{figure}[t]
\centering
\includegraphics[trim={7cm 0cm 3cm 0cm}, clip, width=1.2\columnwidth]{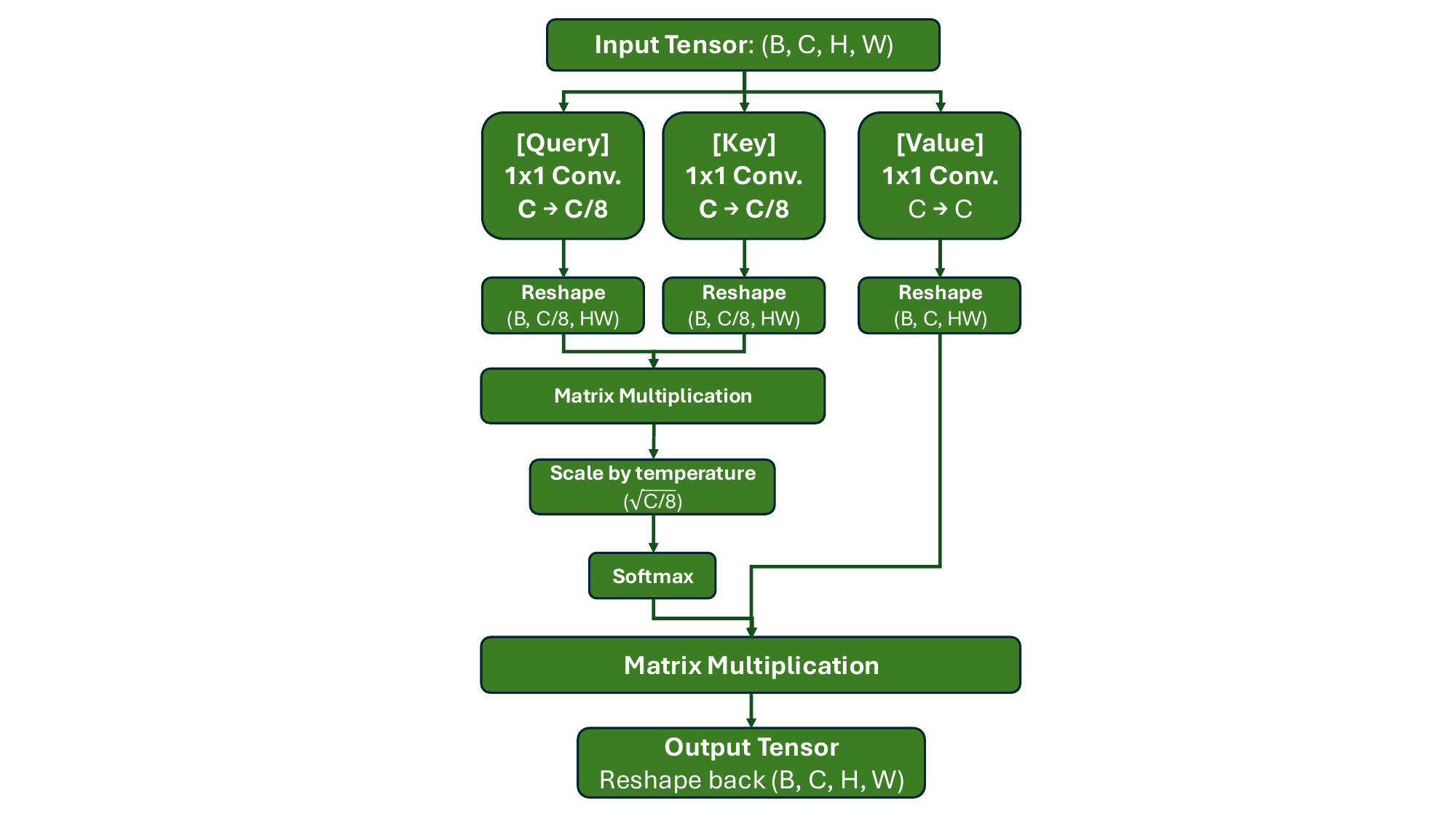}
\caption{Spatio-Temporal Attention Block}
\label{fig:spatiotemporalatt}
\end{figure}

\end{document}